\documentclass[runningheads]{llncs}
\usepackage[misc,geometry]{ifsym}
\usepackage{graphicx}
\usepackage{amsmath}
\usepackage{amssymb}
\usepackage{booktabs}
\usepackage[ruled,vlined]{algorithm2e}
\usepackage{epsfig}
\usepackage{physics}
\usepackage[inline]{enumitem}
\usepackage{comment}
\usepackage{booktabs,framed}
\usepackage{multirow}
\usepackage{xcolor,soul}
\usepackage{latexsym,url}
\usepackage{stackengine} 
\usepackage{wrapfig}

\usepackage[pagebackref,breaklinks]{hyperref}
\usepackage[capitalize]{cleveref}
\crefname{section}{Sec.}{Secs.}
\Crefname{section}{Section}{Sections}
\Crefname{table}{Table}{Tables}
\crefname{table}{Tab.}{Tabs.}

\newcommand{\myfirstpara}[1]{\par \noindent \textbf{#1:}}
\newcommand{\mypara}[1]{ \myfirstpara{#1}}

\DeclareMathOperator*{\argmax}{argmax}

\def\cnn{\texttt{CNN}\xspace}
\def\cnns{\texttt{CNNs}\xspace}
\def\dnn{\texttt{DNN}\xspace}

\def\sota{\texttt{SOTA}\xspace}
\def\gbc{\texttt{GBC}\xspace}
\def\gb{\texttt{GB}\xspace}
\def\us{\texttt{US}\xspace}
\def\wsod{\texttt{WSOD}\xspace}
\def\mil{\texttt{MIL}\xspace}
\def\detr{\texttt{DETR}\xspace}

\def\gbcnet{\texttt{GBCNet}\xspace}

\begin{document}
\title{Gall Bladder Cancer Detection from US Images with Only Image Level Labels}
\titlerunning{Gall Bladder Cancer Detection from US Images}
%
%
\author{
    Soumen Basu \inst{1} \Letter 
    \and
    Ashish Papanai\inst{1} 
    \and
    Mayank Gupta\inst{1} 
    \and
    Pankaj Gupta\inst{1,2}
    \and
    Chetan Arora \inst{1}
}

%
\authorrunning{S. Basu et al.}
%
\institute{
Indian Institute of Technology, Delhi, India \\
\email{soumen.basu@cse.iitd.ac.in}
\and
Postgraduate Institute of Medical Education and Research, Chandigarh, India
}
\maketitle              
%

%
%
\begin{abstract}
	Automated detection of Gallbladder Cancer (GBC) from Ultrasound (US) images is an important problem, which has drawn increased interest from researchers. However, most of these works use difficult-to-acquire information such as bounding box annotations or additional US videos. In this paper, we focus on GBC detection using only image-level labels. Such annotation is usually available based on the diagnostic report of a patient, and do not require additional annotation effort from the physicians. However, our analysis reveals that it is difficult to train a standard image classification model for GBC detection. This is due to the low inter-class variance (a malignant region usually occupies only a small portion of a US image), high intra-class variance (due to the US sensor capturing a 2D slice of a 3D object leading to large viewpoint variations), and low training data availability. We posit that even when we have only the image level label, still formulating the problem as object detection (with bounding box output) helps a deep neural network (DNN) model focus on the relevant region of interest. Since no bounding box annotations is available for training, we pose the problem as weakly supervised object detection (WSOD). Motivated by the recent success of transformer models in object detection, we train one such model, DETR, using multi-instance-learning (MIL) with self-supervised instance selection to suit the WSOD task. Our proposed method demonstrates an improvement of AP and detection sensitivity over the SOTA transformer-based and CNN-based WSOD methods. Project page is at \url{https://gbc-iitd.github.io/wsod-gbc}.
 
\end{abstract}

\begin{figure}[t]
    \centering
    \includegraphics[width=0.9\linewidth]{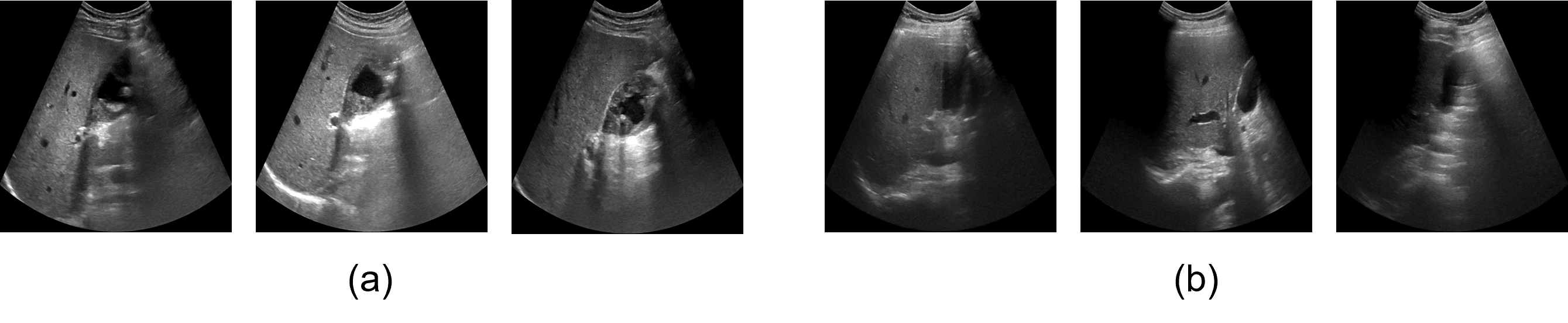}
    \caption{(a) Low inter-class variability. The first two GBs show benign wall thickening, and the third one shows malignant thickening. However, the appearance of the GB in all three images is very similar. (b) High intra-class variability. All three images have been scanned from the same patient, but due to the sensor's scanning plane, the appearances change drastically.}
    \label{fig:teaser}
\end{figure}

%
%
\section{Introduction}
\gbc is a deadly disease that is difficult to detect at an early stage \cite{howlader2017seer,gupta2021locally}. Early diagnosis can significantly improve the survival rate \cite{hong2014surgical}. Non-ionizing radiation, low cost, and accessibility make \us a popular non-invasive diagnostic modality for patients with suspected gall bladder (\gb) afflictions. However, identifying signs of \gbc from routine \us imaging is challenging for radiologists \cite{gupta2020imaging}. In recent years, automated \gbc detection from \us images has drawn increased interest \cite{basu2022surpassing,basu2022unsupervised} due to its potential for improving diagnosis and treatment outcomes. Many of these works formulate the problem as an object detection, since training a image classification model for \gbc detection seems challenging due to the reasons outlined in the abstract (also see \cref{fig:teaser}).

Recently, \gbcnet \cite{basu2022surpassing}, a \cnn-based model, achieved \sota performance on classifying malignant \gb from \us images. \gbcnet uses a two-stage pipeline consisting of object detection followed by classification, and requires bounding box annotations for \gb as well as malignant regions for training. Such bounding box annotations surrounding the pathological regions are time-consuming and require an expert radiologist for annotation. This makes it expensive and non-viable for curating large datasets for training large \dnn models. In another recent work, \cite{basu2022unsupervised} has exploited additional unlabeled video data for learning good representations for downstream \gbc classification and obtained performance similar to \cite{basu2022surpassing} using a ResNet50 \cite{resnet} classifier. The reliance of both \sota techniques on additional annotations or data, limits their applicability. On the other hand, the image-level malignancy label is usually available at a low cost, as it can be obtained readily from the diagnostic report of a patient without additional effort from clinicians. 

Instead of training a classification pipeline, we propose to solve an object detection problem, which involves predicting a bounding box for the malignancy. The motivation is that, running a classifier on a focused attention/ proposal region in an object detection pipeline would help tackle the low inter-class and high intra-class variations. However, since we only have image-level labels available, we formulate the problem as a Weakly Supervised Object Detection (\wsod) problem. As transformers are increasingly outshining \cnns due to their ability to aggregate focused cues from a large area \cite{vit,detr}, we choose to use transformers in our model. However, in our initial experiments \sota \wsod methods for transformers failed miserably. These methods primarily rely on training a classification pipeline and later generating activation heatmaps using attention and drawing a bounding box circumscribing the heatmaps \cite{tscam,scm} to show localization. However, for \gbc detection, this line of work is not helpful as we discussed earlier. 

Inspired by the success of the Multiple Instance Learning (\mil) paradigm for weakly supervised training on medical imaging tasks \cite{transmil,swinmil}, we train a detection transformer, \detr, using the \mil paradigm for weakly supervised malignant region detection. In this, one generates region proposals for images, and then considers the images as bags and region proposals as instances to solve the instance classification (object detection) under the \mil constraints \cite{dietterich1997solving}. At inference, we use the predicted instance labels to predict the bag labels. Our experiments validate the utility of this approach in circumventing the challenges in \us images and detecting \gbc accurately from \us images using only image-level labels.

\mypara{Contributions} The key contributions of this work are:
\begin{itemize}
\itemsep0em
	\item  We design a novel \detr variant based on \mil with self-supervised instance learning towards the weakly supervised disease detection and localization task in medical images. Although \mil and self-supervised instance learning has been used for \cnns \cite{oicr}, such a pipeline has not been used for transformer-based detection models.  
    \item We formulate the \gbc classification problem as a weakly supervised object detection problem to mitigate the effect of low inter-class and large intra-class variances, and solve the difficult \gbc detection problem on \us images without using the costly and difficult to obtain additional annotation (bounding box) or video data.
	\item Our method provides a strong baseline for weakly supervised \gbc detection and localization in \us images, which has not been tackled earlier. Further, to assess the generality of our method, we apply our method to Polyp detection from Colonoscopy images.
\end{itemize}

%
%
\begin{figure}[t]
    \centering
    \includegraphics[width=0.9\textwidth]{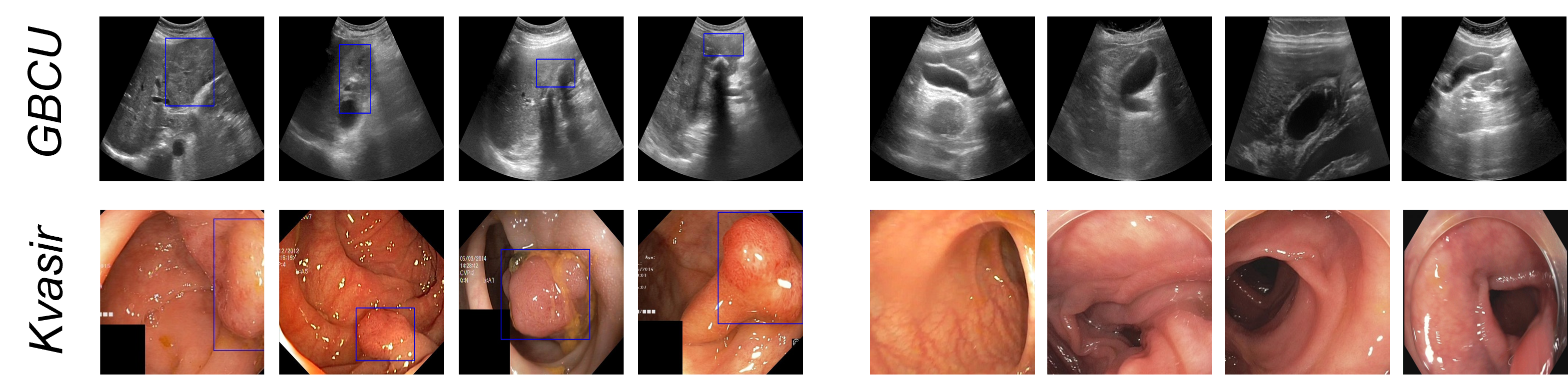}
    \caption{Samples from the GBCU \cite{basu2022surpassing} and Kvasir-SEG \cite{kvasir} datasets. Four images from each of the disease and non-disease classes are shown on the left and right, respectively. Disease locations are shown by drawing bounding boxes.}
    \label{fig:data_sample}
\end{figure}

\section{Datasets}

\myfirstpara{Gallbladder Cancer Detection in Ultrasound Images}
We use the public \gbc \us dataset \cite{basu2022surpassing} consisting of 1255 image samples from 218 patients. The dataset contains 990 non-malignant (171 patients) and 265 malignant (47 patients) \gb images (see \cref{fig:data_sample} for some sample images). The dataset contains image labels as well as bounding box annotations showing the malignant regions. Note that, we use only the image labels for training. We report results on 5-fold cross-validation. We did the cross-validation splits at the patient level, and all images of any patient appeared either in the train or validation split. 

\mypara{Polyp Detection in Colonoscopy Images}
We use the publicly available Kvasir-SEG \cite{kvasir} dataset consisting of 1000 white light colonoscopy images showing polyps (c.f. \cref{fig:data_sample}). Since Kvasir-SEG does not contain any control images, we add 600 non-polyp images randomly sampled from the PolypGen \cite{polypgen} dataset. Since the patient information is not available with the data, we use random stratified splitting for 5-fold cross-validation.

%
%
\section{Our Method}

\begin{figure}[t]
    \centering
    \includegraphics[width=0.9\linewidth]{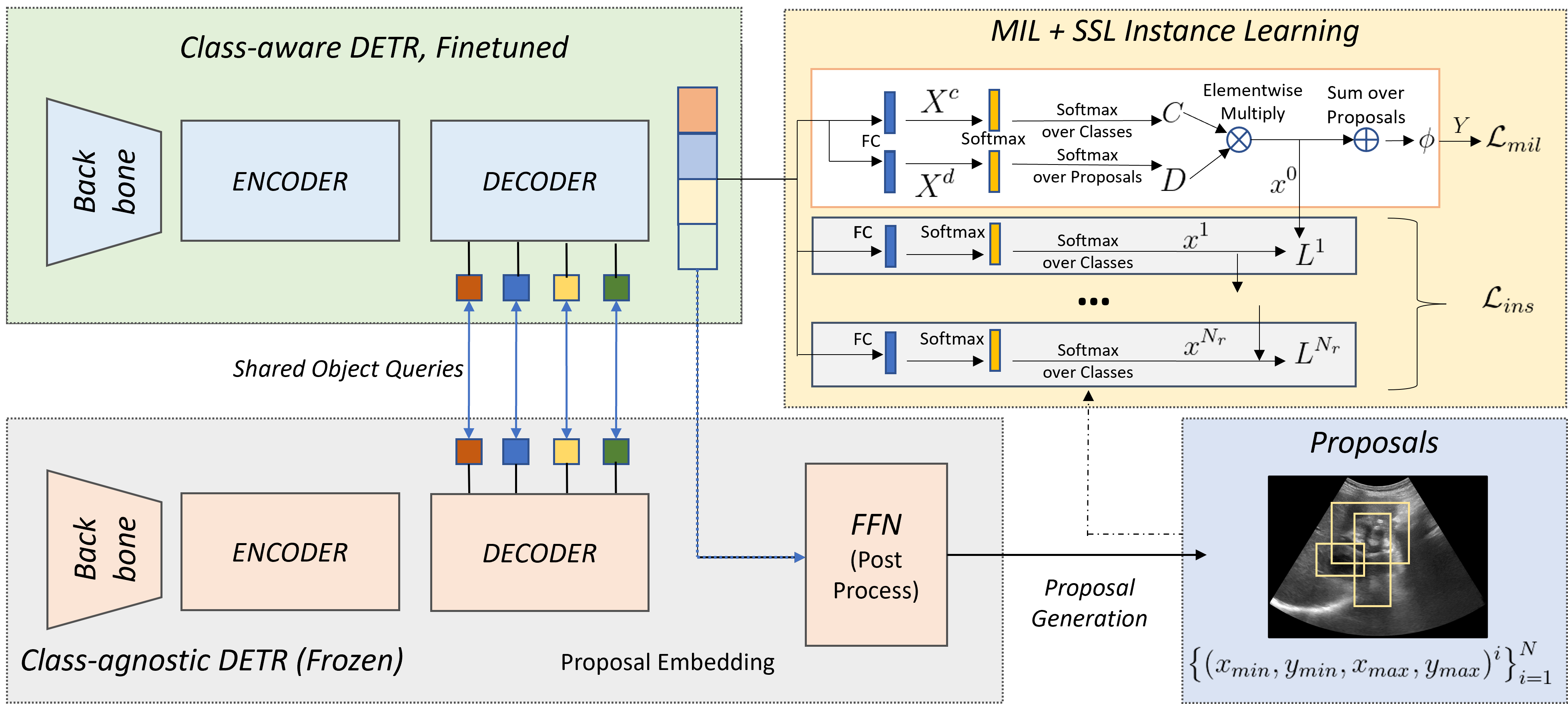}
    \caption{Overview of the proposed Weakly Supervised \detr architecture. The location information in the object queries learned by the class-agnostic \detr ensures generation of high-quality proposals. The \mil framework uses the proposal embeddings generated at the class-aware branch. }
    \label{fig:method}
\end{figure}

\mypara{Revisiting \detr}
The \detr \cite{detr} architectures utilize a \texttt{ResNet} \cite{resnet} backbone to extract \texttt{2D} convolutional features, which are then flattened and added with a positional encoding, and fed to the self-attention-based transformer encoder. The decoder uses cross-attention between learned object queries containing positional embedding, and encoder output to produce output embedding containing the class and localization information. The number of object queries, and the decoder output embeddings is set to 100 in \detr. Subsequently, a feed-forward network generates predictions for object bounding boxes with their corresponding labels and confidence scores. 

\mypara{Proposed Architecture}
\cref{fig:method} gives an overview of our method. We use a \texttt{COCO} pre-trained class-agnostic \detr as proposal generator. The learned object queries contain the embedded positional information of the proposal. Class-agnostic indicates that all object categories are considered as a single object class, as we are only interested in the object proposals. We then finetune a regular, class-aware \detr for the \wsod task. This class-aware \detr is initialized with the checkpoint of the class-agnostic \detr. The learned object queries from the class-agnostic \detr is frozen and shared with the \wsod \detr during finetuning to ensure that the class-aware \detr attends similar locations of the object proposals. The class-agnostic \detr branch is frozen during the finetuning phase. We finally use the \mil-based instance classification with the self-supervised instance learning over the finetuning branch. For \gbc classification, if the model generates bounding boxes for the input image, then we predict the image to be malignant, since the only object present in the data is the cancer. 

\mypara{\mil Setup}
The decoder of the fine-tuning \detr generates $R$ $d$-dimensional output embeddings. Each embedding corresponds to a proposal generated by the class-agnostic \detr. We pass these embeddings as input to two branches with \texttt{FC} layers to obtain the matrices $X^c \in \mathbb{R}^{R\times N_c}$ and $X^r \in \mathbb{R}^{R\times N_c}$, where $R$ is the number of object queries (same as proposals) and $N_c$ is the number of object (disease) categories. Let $\sigma(\cdot)$ denote the softmax operation. 
We then generate the class-wise and detection-wise softmax matrices $C\in\mathbb{R}^{R\times N_c}$ and $D\in\mathbb{R}^{R\times N_c}$, where $C_{ij} = \sigma((X^c)^T_j)i$ and $D_{ij} = \sigma(X^r_i)j$, and $X_i$ denotes the $i$-th row of $X$. $C$ provides classification probabilities of each proposal, and $D$ provides the relative score of the proposals corresponding to each class. The two matrices are element-wise multiplied and summed over the proposal dimension to generate the image-level classification predictions, $\phi\in\mathbb{R}^{N_c}$: 
\begin{align}
\phi_j = \sum_{i=1}^R C_{ij}\cdot D_{ij}
\end{align}
Notice, $\phi_j \in (0,1)$ since $C_{ij}$ and $D_{ij}$ are normalized. Finally, the negative log-likelihood loss between the predicted labels, and image labels $y\in\mathbb{R}^{N_c}$ is computed as the \mil loss:
\begin{align}
\mathcal{L}_\text{mil} = -\sum_{i=1}^{N_c}[{y_i \log{\phi_i}} + (1-y_i) \log{(1-\phi_i)}]   
\end{align}
The \mil classifier further suffers from overfitting to the distinctive classification features due to the mismatch of classification and detection probabilities \cite{oicr}. To tackle this, we further use a self-supervised module to improve the instances.

\mypara{Self-supervised Instance Learning}
Inspired by \cite{oicr}, we design a instance learning module with $N_r$ blocks in a self-supervised framework to refine the instance scores with instance-level supervision. Each block consists of an \texttt{FC} layer. A class-wise softmax is used to generate instance scores $x^n \in \mathbb{R}^{R\times(N_c+1)}$ at $n$-th block. $N_c+1$ includes the background/ no-finding class. Instance supervision of each layer ($n$) is obtained from the scores of the previous layer ($x^{(n-1)}$). The instance supervision for the first layer is obtained from the \mil head. Suppose $\hat{y}^n \in \mathbb{R}^{R\times(N_c+1)}$ is the pseudo-labels of the instances. An instance ($p_j$) is labelled 1 if it overlaps with the highest-scoring instance by a chosen threshold. Otherwise, the instance is labeled $0$ as defined in \cref{eqn:inst_sample}:
\begin{align}
    m^n_j = \argmax_{i}{x^{(n-1)}_{ij}} ~;
    \qquad
    \hat{y}^n_{ij} =  
    \begin{cases}
    1, & IoU(p_j, p_{m^n_j}) \geq \tau\\
    0, & \text{otherwise}
    \end{cases}
    \label{eqn:inst_sample}
\end{align}

The loss over the instances is given by \cref{eqn:inst_loss}:
\begin{align}
    \mathcal{L}_{ins} = - \frac{1}{N_r} \sum_{n=1}^{N_r} \frac{1}{R} \sum_{i=1}^R \sum_{j=1}^{N_c+1} w^{n}_{i} \hat{y}^{n}_{ij} \log x^n_{ij}
    \label{eqn:inst_loss}
\end{align}
Here $x^n_{ij}$ denotes the score of $i$-th instance for $j$-th class at layer $n$. Following \cite{oicr}, the loss weight $w^n_{i} = x^{(n-1)}_{i\,m^n_j}$ is applied to stabilize the loss.
Assuming $\lambda$ to be a scaling value, the overall loss function is given in \cref{eqn:loss}:
\begin{align}
    \mathcal{L} = \mathcal{L}_{mil} + \lambda\mathcal{L}_{ins} 
    \label{eqn:loss}
\end{align}

%
%
\section{Experiments and Results}

\begin{table}[t]
	\centering
	\setlength{\tabcolsep}{10pt}
	\caption{Weakly supervised disease detection performance comparison of our method and SOTA baselines in GBC and Polyps. We report Average Precision at IoU 0.25 ($AP_{25}$).}
	\begin{tabular}{lcc}
		\toprule
		\multirow{2}{*}{\textbf{Method}} & \textbf{GBC} & \textbf{Polyp} \\ 
        & $AP_{25}$ & $AP_{25}$ \\
        \midrule
        TS-CAM \cite{tscam} (ICCV 2021) & 0.024 $\pm$ 0.008 & 0.058 $\pm$ 0.015 \\
        SCM \cite{scm} (ECCV 2022) & 0.013 $\pm$ 0.001 & 0.082 $\pm$ 0.036 \\
        OD-WSCL \cite{odwscl} (ECCV 2022) & 0.482 $\pm$ 0.067 & 0.239 $\pm$ 0.032 \\
        WS-DETR \cite{wsdetr} (WACV 2023) & 0.520 $\pm$ 0.088 & 0.246 $\pm$ 0.023 \\
        Point-Beyond-Class \cite{pointdetr} (MICCAI 2022) & 0.531 $\pm$ 0.070 & 0.283 $\pm$ 0.022 \\
        \midrule
        Ours & 0.628 $\pm$ 0.080 & 0.363 $\pm$ 0.052 \\
        \bottomrule
    \end{tabular}
    \label{tab:wsod}
\end{table}

\begin{figure}[t]
    \centering
    \includegraphics[width=0.95\textwidth]{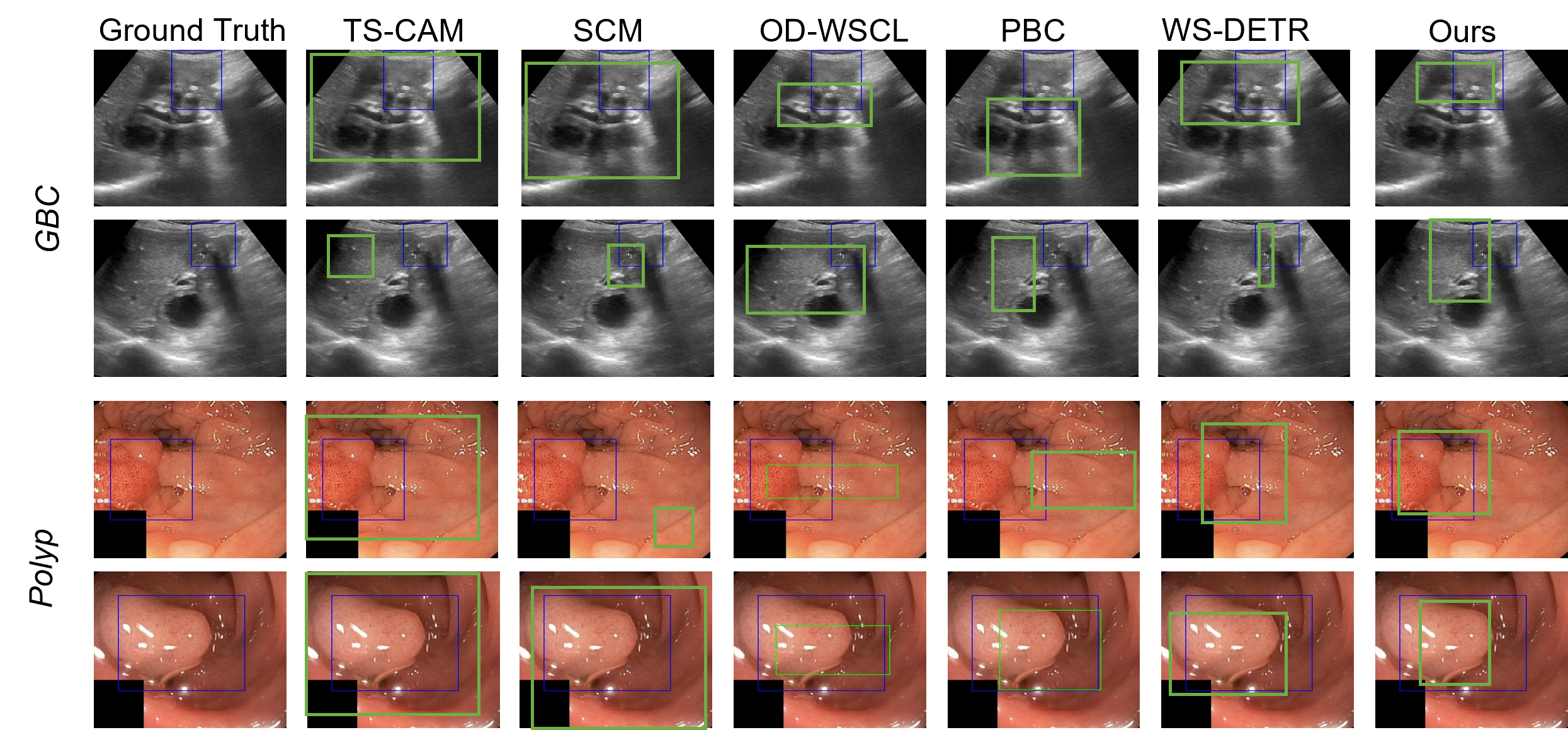}
    \caption{Qualitative analysis of the predicted bounding boxes. Ground truths are in blue, and predictions are in green. We compare with \sota \wsod techniques and our proposed method. Our method predicts much tighter bounding boxes that cover the clinically significant disease regions.}
    \label{fig:visuals}
\end{figure}
\begin{table}[t]
	\centering
	\setlength{\tabcolsep}{6pt}
	\caption{Ablation study. Performance of \mil-framework variants on \detr. We compare the AP and detection sensitivity.} 
	\resizebox{ \linewidth}{!}{%
	\begin{tabular}{lcccc}
		\toprule
        \multirow{2}{*}{\textbf{Design}} & \multicolumn{2}{c}{\textbf{GBC}} & \multicolumn{2}{c}{\textbf{Polyp}} \\
		 & \textbf{$AP_{25}$} & \textbf{Sens.} & \textbf{$AP_{25}$} & \textbf{Sens.} \\
		\midrule
            MIL + DETR & 0.520 $\pm$ 0.088 & 0.833 $\pm$ 0.034 & 0.246 $\pm$ 0.023 & 0.882 $\pm$ 0.034\\	        
            MIL + SSL + DETR (Ours) & 0.628 $\pm$ 0.080 & 0.861 $\pm$ 0.089 & 0.363 $\pm$ 0.052 & 0.932 $\pm$ 0.022\\
		\bottomrule
	\end{tabular}
	}
	\label{tab:ablation}
\end{table}

\begin{table}[t]
	\centering
    \setlength{\tabcolsep}{6pt}
 	\caption{Performance comparison of our method and other \sota methods in \gbc classification. We report accuracy, specificity, and sensitivity. }
	\resizebox{ \linewidth}{!}{%
	\begin{tabular}{llccc}
		\toprule
		{\textbf{Type}} & {\textbf{Method}} & {\textbf{Acc.}} &  {\textbf{Spec.}} & {\textbf{Sens.}} \\
		\midrule
		\multirow{2}{*}{CNN Classifier} 
		& ResNet50 \cite{resnet} & 0.867 $\pm$ 0.031 & 0.926 $\pm$ 0.069 & 0.672 $\pm$ 0.147  \\
		& InceptionV3 \cite{inception} & 0.869 $\pm$ 0.039 & 0.913 $\pm$ 0.032 & 0.708 $\pm$ 0.078 \\
        \midrule
        \multirow{4}{*}{Transformer Classifier} 
        & ViT \cite{vit} & 0.803 $\pm$ 0.078 & 0.901 $\pm$ 0.050 & 0.860 $\pm$ 0.068  \\
		& DEIT \cite{touvron2021training} & 0.829 $\pm$ 0.030 & 0.900 $\pm$ 0.040 & 0.875 $\pm$ 0.063 \\
		& PVTv2 \cite{wang2021pvtv2} & 0.824 $\pm$ 0.033 & 0.887 $\pm$ 0.057 & 0.894 $\pm$ 0.076 \\
        & RadFormer \cite{basu2023radformer} & 0.921 $\pm$ 0.062 & 0.961 $\pm$ 0.049 & 0.923 $\pm$ 0.062  \\
		\midrule
        \multirow{4}{*}{Additional Data/ Annotation}
		& USCL \cite{uscl} & 0.889 $\pm$ 0.047 & 0.895 $\pm$ 0.054 & 0.869 $\pm$ 0.097 \\
        & US-UCL \cite{basu2022unsupervised} & 0.920 $\pm$ 0.034 & 0.926 $\pm$ 0.043 & 0.900 $\pm$ 0.046  \\
		& GBCNet \cite{basu2022surpassing} & 0.921 $\pm$ 0.029 & 0.967 $\pm$ 0.023 & 0.919 $\pm$ 0.063 \\
        & Point-Beyond-Class \cite{pointdetr} & 0.929  $\pm$  0.013 & 0.983  $\pm$  0.042 & 0.731  $\pm$  0.077 \\
        \midrule
		\multirow{4}{*}{SOTA WSOD} &
		TS-CAM \cite{tscam} & 0.862 $\pm$ 0.049 & 0.879 $\pm$ 0.049 & 0.751 $\pm$ 0.045   \\
		& SCM \cite{scm} & 0.795 $\pm$ 0.101 & 0.783 $\pm$ 0.130 & 0.849 $\pm$ 0.072  \\
		& OD-WSCL \cite{odwscl} & 0.815 $\pm$ 0.144 & 0.805 $\pm$ 0.129 & 0.847 $\pm$ 0.214   \\
        & WS-DETR \cite{wsdetr} & 0.839 $\pm$ 0.042 & 0.843 $\pm$ 0.028 & 0.833 $\pm$ 0.034 \\
		\midrule
		WSOD & Ours & 0.834 $\pm$ 0.057 & 0.817 $\pm$ 0.061 & 0.861 $\pm$ 0.089 \\
		\bottomrule
	\end{tabular}
	}
	\label{tab:key_results}
\end{table}

\begin{table}[t]
	\centering
	\setlength{\tabcolsep}{10pt}
	\caption{Comparison with \sota \wsod baselines in classifying Polyps from Colonoscopy images. }
    \resizebox{ 0.9\linewidth}{!}{%
	\begin{tabular}{lccc}
		\toprule
        \textbf{Method} & \textbf{Acc.} & \textbf{Spec.} & \textbf{Sens.}  \\
		\midrule
		TS-CAM \cite{tscam} & 0.704 $\pm$ 0.017 & 0.394 $\pm$ 0.042 & 0.891 $\pm$ 0.054 \\
		SCM \cite{scm} & 0.751 $\pm$ 0.026 & 0.523 $\pm$ 0.014 & 0.523 $\pm$ 0.016  \\
		OD-WSCL\cite{odwscl} & 0.805 $\pm$ 0.056 & 0.609 $\pm$ 0.076 & 0.923 $\pm$ 0.034  \\
		WS-DETR \cite{wsdetr} & 0.857 $\pm$ 0.071 & 0.812 $\pm$ 0.088 & 0.882 $\pm$ 0.034 \\
		Point-Beyond-Class \cite{pointdetr} & 0.953 $\pm$ 0.007 & 0.993 $\pm$ 0.004 & 0.924 $\pm$ 0.011 \\
		\midrule
		Ours & 0.878 $\pm$ 0.067 & 0.785 $\pm$ 0.102 & 0.932 $\pm$ 0.022 \\
		\bottomrule
	\end{tabular}
	}
	\label{tab:polyp_results}
\end{table}

\myfirstpara{Experimental Setup}
We use a machine with Intel Xeon Gold 5218@2.30GHz processor and 8 Nvidia Tesla V100 GPUs for our experiments. The model is trained using SGD with LR 0.001 (for MIL head), weight decay $10^{-6}$, and momentum 0.9 for 100 epochs with batch size 32. The LR at backbone and transformer are 0.003, and 0.0003, respectively. We use a cosine annealing of the LR. 

\mypara{Comparison with SOTA}
\cref{tab:wsod} shows the bounding box localization results of the \wsod task. Our method surpasses all latest \sota \wsod techniques by 9 points, and establishes itself as a strong \wsod baseline for \gbc localization in \us images. Our method also achieves 7-point higher AP score for polyp detection. We present visualizations of the predicted bounding boxes in \cref{fig:visuals} which shows that the localization by our method is more precise and clinically relevant as compared to the baselines. 

\mypara{Generality of the Method}
We assess the generality of our method by applying it to polyp detection on colonoscopy images. The applicability of our method on two different tasks - (1) \gbc detection from \us and (2) Polyp detection from Colonoscopy, indicates the generality of the method across modalities. 

\mypara{Ablation Study}
We show the detection sensitivity to the self-supervised instance learning module in \cref{tab:ablation} for two variants, (1) vanilla \mil head on \detr, and (2) \mil with self-supervised instance learning on \detr. \cref{tab:ablation} shows the Average Precision and detection sensitivity for both diseases. The results establish the benefit of using the self-supervised instance learning. Other ablations related to the hyper-parameter sensitivity is given in Supplementary Fig. S1.

\mypara{Classification Performance}
We compare our model with the standard \cnn-based and Transformer-based classifiers, \sota \wsod-based classifiers, and \sota classifiers using additional data or annotations (\cref{tab:key_results}). Our method beats the \sota weakly supervised techniques and achieves 1.2\% higher sensitivity for GBC detection. The current \sota \gbc detection models require additional bounding box annotation \cite{basu2022surpassing} or, \us videos \cite{basu2022unsupervised,uscl}. However, even without these additional annotations/ data, our method reaches 86.1\% detection sensitivity. The results for polyp classification are reported in \cref{tab:polyp_results}. Although our method has a slightly lower specificity, the sensitivity surpasses the baselines reported in literature \cite{jha2021real}, and the \sota \wsod based baselines. 

%
%
\section{Conclusion}
\gbc is a difficult-to-detect disease that benefits greatly from early diagnosis. 
While automated \gbc detection from \us images has gained increasing interest from researchers, training a standard image classification model for this task is challenging due to the low inter-class variance and high intra-class variability of malignant regions. 
Current \sota models for \gbc detection require costly bounding box annotation of the pathological regions, or additional \us video data, which limit their applicability. We proposed to formulate \gbc detection as a weakly supervised object detection/ localization problem using a \detr with self-supervised instance learning in a \mil framework. Our experiments show that the approach achieves competitive performance without requiring additional annotation or data. We hope that our technique will simplify the model training at the hospitals with easily available data locally, enhancing the applicability and impact of automated \gbc detection. 

%
%
\bibliographystyle{splncs04}
\bibliography{reference}

\begin{thebibliography}{10}
\providecommand{\url}[1]{\texttt{#1}}
\providecommand{\urlprefix}{URL }
\providecommand{\doi}[1]{https://doi.org/#1}

\bibitem{polypgen}
Ali, S., Jha, D., Ghatwary, N., Realdon, S., Cannizzaro, R., Salem, O.E.,
  Lamarque, D., Daul, C., Riegler, M.A., Anonsen, K.V., et~al.: a multi-centre
  polyp detection and segmentation dataset for generalisability assessment.
  Scientific Data  \textbf{10}(1), ~75 (2023)

\bibitem{scm}
Bai, H., Zhang, R., Wang, J., Wan, X.: Weakly supervised object localization
  via transformer with implicit spatial calibration. In: ECCV. pp. 612--628.
  Springer (2022)

\bibitem{basu2022surpassing}
Basu, S., Gupta, M., Rana, P., Gupta, P., Arora, C.: Surpassing the human
  accuracy: Detecting gallbladder cancer from usg images with curriculum
  learning. In: CVPR. pp. 20886--20896 (2022)

\bibitem{basu2023radformer}
Basu, S., Gupta, M., Rana, P., Gupta, P., Arora, C.: Radformer: Transformers
  with global--local attention for interpretable and accurate gallbladder
  cancer detection. Medical Image Analysis  \textbf{83},  102676 (2023)

\bibitem{basu2022unsupervised}
Basu, S., Singla, S., Gupta, M., Rana, P., Gupta, P., Arora, C.: Unsupervised
  contrastive learning of image representations from ultrasound videos with
  hard negative mining. In: MICCAI. pp. 423--433. Springer (2022)

\bibitem{detr}
Carion, N., Massa, F., Synnaeve, G., Usunier, N., Kirillov, A., Zagoruyko, S.:
  End-to-end object detection with transformers. In: ECCV. pp. 213--229.
  Springer (2020)

\bibitem{uscl}
Chen, Y., Zhang, C., Liu, L., Feng, C., Dong, C., Luo, Y., Wan, X.: Uscl:
  Pretraining deep ultrasound image diagnosis model through video contrastive
  representation learning. In: MICCAI. pp. 627--637. Springer (2021)

\bibitem{dietterich1997solving}
Dietterich, T.G., Lathrop, R.H., Lozano-P{\'e}rez, T.: Solving the multiple
  instance problem with axis-parallel rectangles. Artificial intelligence
  \textbf{89}(1-2),  31--71 (1997)

\bibitem{vit}
Dosovitskiy, A., Beyer, L., Kolesnikov, A., Weissenborn, D., Zhai, X.,
  Unterthiner, T., Dehghani, M., Minderer, M., Heigold, G., Gelly, S., et~al.:
  An image is worth 16x16 words: Transformers for image recognition at scale.
  arXiv preprint arXiv:2010.11929  (2020)

\bibitem{tscam}
Gao, W., Wan, F., Pan, X., Peng, Z., Tian, Q., Han, Z., Zhou, B., Ye, Q.:
  Ts-cam: Token semantic coupled attention map for weakly supervised object
  localization. In: ICCV. pp. 2886--2895 (2021)

\bibitem{gupta2020imaging}
Gupta, P., Marodia, Y., Bansal, A., Kalra, N., Kumar-M, P., Sharma, V., Dutta,
  U., Sandhu, M.S.: Imaging-based algorithmic approach to gallbladder wall
  thickening. World journal of gastroenterology  \textbf{26}(40), ~6163 (2020)

\bibitem{gupta2021locally}
Gupta, P., Meghashyam, K., Marodia, Y., Gupta, V., Basher, R., Das, C.K.,
  Yadav, T.D., Irrinki, S., Nada, R., Dutta, U.: Locally advanced gallbladder
  cancer: a review of the criteria and role of imaging. Abdominal Radiology
  \textbf{46}(3),  998--1007 (2021)

\bibitem{resnet}
He, K., Zhang, X., Ren, S., Sun, J.: Deep residual learning for image
  recognition. In: CVPR. pp. 770--778 (2016)

\bibitem{hong2014surgical}
Hong, E.K., Kim, K.K., Lee, J.N., Lee, W.K., Chung, M., Kim, Y.S., Park, Y.H.:
  Surgical outcome and prognostic factors in patients with gallbladder
  carcinoma. Annals of Hepato-Biliary-Pancreatic Surgery  \textbf{18}(4),
  129--137 (2014)

\bibitem{howlader2017seer}
Howlader, N., Noone, A., Krapcho, M., Miller, D., Bishop, K., Kosary, C., Yu,
  M., Ruhl, J., Tatalovich, Z., Mariotto, A., et~al.: Seer cancer statistics
  review, 1975-2014, national cancer institute. Bethesda, MD pp. 1--12 (2017)

\bibitem{jha2021real}
Jha, D., Ali, S., Tomar, N.K., Johansen, H.D., Johansen, D., Rittscher, J.,
  Riegler, M.A., Halvorsen, P.: Real-time polyp detection, localization and
  segmentation in colonoscopy using deep learning. IEEE Access  \textbf{9},
  40496--40510 (2021)

\bibitem{kvasir}
Jha, D., Smedsrud, P.H., Riegler, M.A., Halvorsen, P., de~Lange, T., Johansen,
  D., Johansen, H.D.: Kvasir-seg: A segmented polyp dataset. In: MMM. pp.
  451--462. Springer (2020)

\bibitem{pointdetr}
Ji, H., Liu, H., Li, Y., Xie, J., He, N., Huang, Y., Wei, D., Chen, X., Shen,
  L., Zheng, Y.: Point beyond class: A benchmark for weakly semi-supervised
  abnormality localization in chest x-rays. In: MICCAI. pp. 249--260. Springer
  (2022)

\bibitem{wsdetr}
LaBonte, T., Song, Y., Wang, X., Vineet, V., Joshi, N.: Scaling novel object
  detection with weakly supervised detection transformers. In: WACV. pp. 85--96
  (2023)

\bibitem{swinmil}
Qian, Z., Li, K., Lai, M., Chang, E.I.C., Wei, B., Fan, Y., Xu, Y.: Transformer
  based multiple instance learning for weakly supervised histopathology image
  segmentation. In: MICCAI. pp. 160--170. Springer Nature Switzerland Cham
  (2022)

\bibitem{odwscl}
Seo, J., Bae, W., Sutherland, D.J., Noh, J., Kim, D.: Object discovery via
  contrastive learning for weakly supervised object detection. In: ECCV. pp.
  312--329. Springer (2022)

\bibitem{transmil}
Shao, Z., Bian, H., Chen, Y., Wang, Y., Zhang, J., Ji, X., et~al.: Transmil:
  Transformer based correlated multiple instance learning for whole slide image
  classification. NeurIPS  \textbf{34},  2136--2147 (2021)

\bibitem{inception}
Szegedy, C., Vanhoucke, V., Ioffe, S., Shlens, J., Wojna, Z.: Rethinking the
  inception architecture for computer vision. In: Proceedings of the IEEE
  conference on computer vision and pattern recognition. pp. 2818--2826 (2016)

\bibitem{oicr}
Tang, P., Wang, X., Bai, X., Liu, W.: Multiple instance detection network with
  online instance classifier refinement. In: CVPR. pp. 2843--2851 (2017)

\bibitem{touvron2021training}
Touvron, H., Cord, M., Douze, M., Massa, F., Sablayrolles, A., J{\'e}gou, H.:
  Training data-efficient image transformers \& distillation through attention.
  In: ICML. pp. 10347--10357. PMLR (2021)

\bibitem{wang2021pvtv2}
Wang, W., Xie, E., Li, X., Fan, D.P., Song, K., Liang, D., Lu, T., Luo, P.,
  Shao, L.: Pvtv2: Improved baselines with pyramid vision transformer (2021)

\end{thebibliography}

\end{document}